\documentclass{IOS-Book-Article}

\def\hb{\hbox to 10.7 cm{}}
\usepackage[show]{ed} % hide for final version

\usepackage[utf8]{inputenc}

\usepackage{lipsum}

\usepackage{lstsemantic}
\usepackage[T1]{fontenc}
\usepackage[scaled=.9]{beramono}
\usepackage{graphicx}
\usepackage{subfigure}
\usepackage{macros}
\usepackage{amssymb}
\usepackage{tikz}
\usepackage{url}
\usepackage{xspace}
\usepackage{float}
\usepackage{multicol}

\newcommand{\technical}[1]{\textsf{#1}}

\newcommand{\GODP}[1]{\technical{{#1}}}

\newcommand{{\DOL}}{\textmd{\textsc{Dol}}\xspace}
\newcommand{{\CASL}}{\textmd{\textsc{Casl}}\xspace}
\newcommand{{\Hets}}{\textmd{\textsc{Hets}}\xspace}

\newcommand{{{\GenericDOL}}}{{\textmd{\textsc{Generic Dol}}}\xspace}
\newcommand{\GDOL}{\textmd{\textsc{GDol}}\xspace}
\newcommand{{\GenericODP}}{\textmd{\textsc{Generic Ontology Development Pattern}}\xspace}

\newcommand{\class}[1]{\technical{#1}}
\newcommand{\relation}[1]{\technical{{#1}}}

\newcommand{\Keyword}[1]{\technical{{\textbf{#1}}}}

\newcommand{\technicalterm}[1]{\textit{#1}}

\newcommand{\seecite}[1]{\cite{#1}}

\begin{document}

	\pagestyle{headings}
	\def\thepage{}
	
\begin{frontmatter}

%\pretitle{Pretitle}
\title{%Generic Ontologies and \\
        Generic Ontology Design Patterns \\ %for Ontology Development
        at Work
        }
\markboth{}{June 2019\hb}
%\subtitle{Subtitle}

\author[A]{\fnms{Bernd} \snm{Krieg-Br\"uckner}%
% \thanks{Corresponding Author: Book Production Manager, IOS Press, Nieuwe Hemweg 6B,
% 1013 BG Amsterdam, The Netherlands; E-mail:
% bookproduction@iospress.nl.}
},
\author[B]{\fnms{Till} \snm{Mossakowski}}
and
\author[B]{\fnms{Fabian} \snm{Neuhaus}}

\runningauthor{B.~Krieg-Br\"uckner, T.~Mossakowski, F.~Neuhaus}
\address[A]{
Collaborative Research Center EASE,
Universit\"at Bremen, Germany}
\address[B]{
Institute for Intelligent Cooperating Systems,
Faculty of Computer Science,
Otto-von-Guericke-Universit\"at Magdeburg, Germany
}

\begin{abstract}
%\technicalterm{Generic Ontologies}
\technicalterm{Generic Ontology Design Pattern}s, GODPs, are defined in {{\GenericDOL}},
%are introduced in \technicalterm{GDOL},
an extension of %\technicalterm{DOL},
the  \technicalterm{Distributed Ontology, Model and Specification Language},
and implemented using %{\Hets},
the \technicalterm{Heterogeneous Tool Set}.
Parameters such as classes, properties, individuals, or whole ontologies
may be instantiated with arguments in a host ontology.
%A \technicalterm{Generic Ontology Design Pattern}, GODP, defined in GDOL,
%embodies an ontology development operation and serves as a metho\-do\-logical tool for %safer
%ontology development.
%Some non-trivial GODPs are presented for illustration,
%e.g. for %domain independent

The potential of {{\GenericDOL}}
is illustrated with GODPs for an example from the literature, namely \GODP{Role}.
We will also discuss how larger GODPs may be composed by instantiating smaller GODPs.
\end{abstract}

\begin{keyword}
ontology design patterns 
\end{keyword}
\end{frontmatter}

%
% \lstset{% general command to set parameter(s)
% basicstyle=\small, % print whole listing small
% basicstyle=\ttfamily,
% %keywordstyle=\color{black}\bfseries\underbar, % underlined bold black keywords
% %identifierstyle=, % nothing happens
% %commentstyle=, % white comments
% stringstyle=\sffamily,
% showstringspaces=false,
% %language={},
% numbers=left, numberstyle=\tiny, stepnumber=2, numbersep=5pt,
% language=dolText,alsolanguage=owl2Manchester}

%+++++++++++++++++++++++ Introduction ++++++++++++++++++++++++++

\section{Introduction}

\label{sec:introduction}

%\subsection{Paradigm Shift in Ontology Engineering}
%\paragraph{Paradigm shift in ontology engineering.}
\label{sect:ParadigmShift-intro}
In ontology engineering we may distinguish (at least) three kinds of stakeholders:
ontology experts, domain experts and end-users. For each, the kind and level of expertise is quite different: End-users should not be required to have ontology expertise and may have little domain knowledge; domain experts often have little ontology expertise; ontology experts have little specialised domain expertise.

Since many ontology development projects are small and, frequently, not well funded, many of them do not involve a dedicated ontology expert at all or get only very limited input from an ontology expert during the design phase of the ontology. Thus, the majority of the development of an ontology is typically entrusted to domain experts. This may lead to  poor design choices and avoidable errors. Furthermore, because of their lack of experience, domain experts may not be able to identify opportunities to reuse existing best practices and ontologies.

\label{sect:OntologyDesignPatterns-intro}
%Structuring in-the-small is required for sustained confidence in %(better: correctness of)
%intricate local, reusable ontology development tasks.
%There is increasing demand for easier reuse of (bits and pieces or larger chunks of) ontologies.
%
%\technicalterm{Ontology Design Patterns}, ODPs,
Ontology Design Patterns (ODPs) have been proposed for some time as methodology for ontology development, see the early work by \cite{OntologyPatterns_gangemi}, the compilation in \cite{DBLP:books/ios/HGJKP2016}, and the %excellent
review of the state of the art in \cite{OPLA-WOP17}.
In theory, ODPs provide a solution for the lack of ontology experts: ODPs enable domain ontologists to reuse existing best practices and design decisions and, thus, benefit from the experience of ontology experts who developed the ODPs
and lead to a considerable increase in quality of ontologies.
However, in practice the adaptation of ODPs as tools for ontology engineers has been  slow. %cumbersome and error-prone.\ednote{Citation?}
In our opinion this is caused by the fact that currently the utilisation of ODPs is cumbersome for ontology developers.

Let us assume  an ontology developer intends to reuse a given ODP $P$ for the representation of some domain $D$ during the development of an ontology $O$,  there are basically two options. She may import $P$ into $O$, include the relevant terms for $D$ and link these terms via
 subclass and subproperty relationships to $P$.
However,  this approach  clutters the resulting ontology $O$, because $O$ includes not just terms about $D$, but all of the generic terms of $P$. Moreover,  an ODP often does not fit into the context of a given ontology completely, but needs to be adapted and some axioms need to be removed. Importation does not provide a means to change $P$, either one imports it completely into $O$ or not.
Alternatively, the ontology developer may include an axiomatisation of the terms about $D$ that mirrors the axiomatisations of  $P$ structurally. This approach allows more flexibility for adaptions, but manual redesign is error-prone and time-intensive, and the goal of shielding the domain ontologist from the complex design decisions during ontology development is not achieved.

In this paper we propose \technicalterm{Generic Ontology Design Patterns}, GODPs, as a methodology for representing and instantiating ODPs in a way that is adaptable, and allows domain experts (and other users) to safely use ODPs  without cluttering their ontologies.
The main ideas behind GODPs (Sect.\,\ref{sect:GenericOntologiesGDOL}) are the following: ODPs are expressed in a dedicated formal, parametrised pattern language that allows (a) the definition of ODPs, (b) specify instantiations of ODPs, (c) extend, modify, and combine ODPs to larger ODPs, and (d) describe the relationships between ODPs.
\label{sect:GODPs-intro}
%We take generic ontologies as the basis for \technicalterm{Generic Ontology Design Patterns},
GODPs  embody dedicated development operations.
They are defined in an extension of the \technicalterm{Distributed Ontology, Model and Specification Language}, DOL,
and implemented using %a tool framework for heterogeneous development,
the \emph{Heterogeneous Tool Set}, {\Hets} (Sect.\,\ref{sect:GenericOntologiesGDOL}).

GODPs are patterns in the sense that they contain typed variables. The definition of a GODP involves the specification of parameters that need to be provided for the instantiation of a GODP. Possible parameters include symbols, lists of symbols, but also whole ontologies.  The latter
enable to express powerful semantic constraints using corresponding axioms; such requirements act like preconditions for instantiations, guaranteeing more consistency and safety.
% In a pattern, ontology parameters may also function as stubs stating requirements for the incorporation of other patterns (Fig.\,\ref{fig:AgentRole}).

In this paper we will introduce GODPs by first discussing a toy example in Section~\ref{sect:GenericOntologiesGDOL} and  afterward an in-depth discussion of the \GODP{Role} ODP from the literature in Section~\ref{fig:RolePatterns}.
%\ednote{revise when paper is more mature, cite papers}
As we will show, GODPs enable the nested use of ODPs, which reduces code duplication.
Since GODPS are written in an extension of DOL, GDOP developers may utilise the other features of 
DOL to, e.g., to explicitly state logical properties of GODPs, represent competency questions, and definitorial extensions. This is significant since it enables quality control on the level of the GODPs as well as their instantiations.

%During reuse, instantiations provide extra handles for improving consistency, checking arguments against structural requirements stated in the parameters.

Thus, GODPs provide one tool to support the strategy of ontology reuse by  modularisation, which has been proposed  Katsumi and Grüninger \cite{Gruninger2017COLORE,DBLP:journals/ao/KatsumiG18,DBLP:journals/ao/KatsumiG17}. 
GODPs shares many objectives with \technicalterm{Parameterized OTTR Templates} with macro expansion by \cite{OTTR-WOP17}.
The differences in the approaches will be analysed in Sect.\,\ref{sect:FutureWork}. 

\section{Generic Ontology Design Patterns in {{\GenericDOL}}}
\label{sect:GenericOntologiesGDOL}
%\paragraph{Structuring with DOL.}
\label{sect:DOL}
\label{sect:Hets}

The \technicalterm{Distributed Ontology, Model and Specification Language}, DOL,
an OMG standard
\seecite{DOL-OMG,MossakowskiEtAl13d,mossakowski2015distributed},
allows the structured definition of ontologies, using import, union, renaming, module extraction,
and many more. Thus, DOL is not yet another ontology language, but a meta-language, which allows to
define and manipulate ontologies and networks of ontologies.  DOL can be used on top of a variety of languages,
in particular OWL 2. 

The left column in Fig.~\ref{fig:DOLexample} contains a small  DOL file. After %a prefix and 
a logic 
is declared two different ontologies are specified, namely \GODP{Driving} and \GODP{DrivingExtended}.\footnote{In the following we will omit %prefix and 
logic declarations.}
%\ednote{Till: hier sollte zum Beispiel noch die Expansion von DrivingPatternInstance als .omn dazu kommen und erklärt werden.
%DONE.}
%\ednote{ich habe prefix gelöscht da nicht relevant hier und nicht genutzt}
%
Note that the expressions in the ontology declaration are not in DOL, but in some ontology language, in this case in OWL Manchester Syntax. (Other ontology languages are available, e.g. FOL-based syntaxes.)  
\GODP{DrivingExtended} is specified as extension of \GODP{Driving}, which contains an additional domain axiom. The  phrase A  \Keyword{then} B in DOL indicates that all definitions in A are visible in B, where A and B are ontologies in OWL (or some other  ontology language) or instantiations of GODPs.
The semantics of \Keyword{and} joining two instantiations is union of the two ontologies, and intersection of the corresponding axioms.

\begin{figure}
\newlength\Colsep
\setlength\Colsep{0pt}
\noindent\begin{minipage}{\textwidth}
\begin{minipage}[c][4cm][c]{\dimexpr0.5\textwidth-0.5\Colsep\relax}
%    \lstinputlisting[firstline=1]{TimeVaryingEntitities/DOLExample.dol}
\begin{lstlisting}[firstline=3]
logic OWL
ontology Driving = 
 Class: Vehicle 
 ObjectProperty: drives 
  Range: Vehicle

ontology DrivingExtended = 
 Driving 
then 
 Class: Person
 ObjectProperty: drives 
  Domain: Person 
\end{lstlisting}
%   \lstinputlisting[firstline=2]{TimeVaryingEntitities/SimpleRelation.dol}
\end{minipage}\hfill
\begin{minipage}[c][6cm][c]{\dimexpr0.5\textwidth-0.5\Colsep\relax}
\begin{lstlisting}[firstline=2]
pattern SimpleRelationGODP
 [ObjectProperty: p]
 [Class: D][Class: R] =
ObjectProperty: p
 Domain: D Range: R
\end{lstlisting}
\vspace{-1.5em}
\begin{lstlisting}[firstline=2]
ontology DrivingPatternInstance = 
SimpleRelationGODP
 [drives][Person][Vehicle]		
 
\end{lstlisting}
\vspace{-1.5em}
\begin{lstlisting}[firstline=2]
Ontology: <DrivingPatternInstance_Exp>
Class: Person
Class: Vehicle
ObjectProperty: drives
 Domain: Person Range: Vehicle
\end{lstlisting}	
\end{minipage}%
\end{minipage}
   \vspace{-1em}
   \caption{{DOL Example},
    and \GODP{Simple Relation} GODP and its application
    }
    \label{fig:DOLexample}

\end{figure}

We developed  %the language
{{\GenericDOL}}, %\technicalterm{Generic DOL}, GDOL,
which extends DOL by a parameterization mechanism for ontologies.
Generics in {{\GenericDOL}} borrow their semantics from {{\CASL}}'s
\emph{generic specification} mechanism \cite{CaslTCS03,CASL-RM,CASL-UM}.
{Generics}
\label{sect:GenericOntologies}
%Generics
(first introduced in ADA \seecite{%ADA-STD,
ADA79}) are not just macros.
Their most important aspect is that all parameters are fully typed items, and argument types are checked against parameter types.

The syntax for OWL in {{\GenericDOL}} is presently Manchester Syntax, extended by parameterised names. % (cf. Sect.\,\ref{sect:ParameterizedNames}) 
E.g., \GODP{SimpleRelationGODP} in  Fig.~\ref{fig:DOLexample} is a very basic pattern that 
is defined to utilise three symbols as parameters, namely p (of type object property), D and R (both of type class). The body of the ontology specification (the part right of the ``='' symbol) contains an ontology in OWL Manchester Syntax, where p, D, and R occur.\footnote{
Strictly speaking, the ontology in the body of the definition of \GODP{SimpleRelationGODP} is not a legal OWL ontology, since the symbols D and R are not declared. However, in \GDOL  is not necessary to declare symbols that are already typed as parameters. 
} 
 The GODP \GODP{SimpleRelationGODP} may be instantiated by providing suitable arguments. 
 E.g., in the definition of the ontology \GODP{drivePatternInstance} the GODP \GODP{SimpleRelationGODP} 
 is instantiated with `drives', `person', and `vehicle'.

The \emph{Heterogeneous Tool Set}, {\Hets} 
\seecite{DBLP:conf/tacas/MossakowskiML07}, is the implementation basis
for DOL and {\GenericDOL}. {\Hets} is able to compute ontologies that are specified in {\GenericDOL} with structuring operations like ``then'', ``and'', and generics. For this purpose \Hets{} interprets  \DOL terms and expands the instantiation of generics. 
This process is called \emph{flattening of an ontology};  the result is a proper OWL ontology.
The flattening of the ontologies \GODP{drivesExtended}  and \GODP{drivePatternInstance} results in the same ontology; it is shown (in pure Manchester syntax) as \GODP{DrivingPatternInstance\_Exp}
at the end of Fig.~\ref{fig:DOLexample}. 
 
 %\paragraph{Ontology Parameters.}
 \label{sect:OntologyParameters}
 Parameters in  {\GenericDOL} 
 are technically ontologies.
 However, single-symbol parameters are recognized as such; in effect,
 parameter kinds for single-symbol parameters in OWL are Class, Individual, ObjectProperty, or DataProperty.
 In general, a  parameter may be a \emph{complex ontology} which contains axioms that specify specific abstract properties.
 An argument ontology must conform to such a parameter ontology, i.e. the required properties must be satisfied. {\Hets} will take care of generating an appropriate proof obligation, if this cannot be deduced automatically.
 This concept makes {\GenericDOL}, and GODPs,
 %structuring of ontologies in-the-large and in-the-small
 extremely powerful to capture semantic preconditions for instantiation (see examples in \cite{GODPs-WOP17}).
 %\emph{separate semantic concepts} independently, beyond mere modularization and ``object-oriented'' inheritance.

Since ontologies \GODP{drivesExtended},  \GODP{drivePatternInstance} and \GODP{DrivingPatternInstance\_Exp} are just different representations of the same axioms (namely,   declaration of drives, its range, and its domain), what is the benefit of using {{\GenericDOL}}?
 After all,  one can just write the OWL ontology \GODP{DrivingPatternInstance\_Exp} directly without the additional complexity. One benefit of both structuring mechanisms and GODPs is an increased modularity, which enables reusability and avoids code duplication. E.g., by dividing the axioms into two modules it is possible to reuse the axioms in \GODP{Driving} independently from the additional axioms in  \GODP{DrivingExtended}. 
 Similarly, it is possible to instantiate \GODP{SimpleRelationGODP} with different parameters to declare a different object property, its domain and range. If at a latter time one needs to change the axioms of \GODP{Driving}  or \GODP{SimpleRelationGODP}, these changes are propagated to all 
  ontologies that are based on them. An additional benefit of GODPs is the fact that it hides the internal complexity of its axiomatisation. For the instantiation of a pattern the user does not need to understand  the internal structure of the GODP, he just needs to provide the appropriate parameters. 
  
These benefits are minuscule in the case of a  simple ontology like 
 \GODP{DrivingPatternInstance\_Exp} in  Fig.~\ref{fig:DOLexample}. 
 To illustrate of power of structuring mechanisms and GODPs in {\GenericDOL}  we will discuss in the following an ontology design pattern from the literature, namely the \emph{Role Pattern}, see  Section~\ref{sec:rolepattern}.

\section{Role Pattern}

\label{sec:rolepattern}
\begin{figure}
\includegraphics[width=0.9\textwidth]{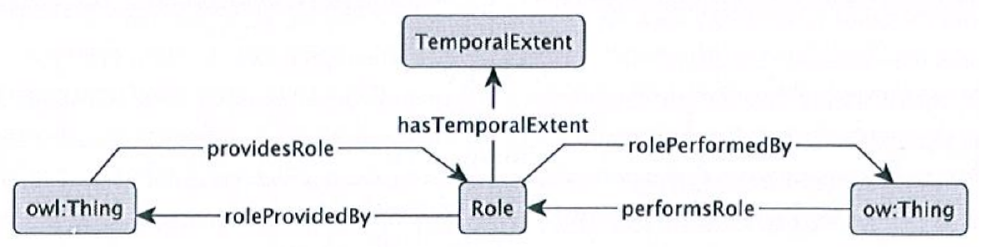}
%\\[1em]
%\includegraphics[width=0.9\textwidth]{figures/AgentRole-diagram.pdf
%}
\vspace{-1em}
\caption{Structure of the Role ODP 
%Role and AgentRole ODPs 
from \cite{DBLP:books/ios/p/Krisnadhi16a}.}
\label{fig:Role-diagram}
\end{figure}

 \begin{figure}[tb]
%   \begin{multicols}{2}
\begin{lstlisting}[firstline=1]
Prefix: : <http://www.ontohub.org/godp/>
Ontology: <Role_Original>
Class: Role 
 SubClassOf: roleProvidedBy max 1 owl:Thing, 
             rolePerformedBy max 1 owl:Thing,
             hasTemporalExtent some TemporalExtent, 
             hasTemporalExtent only TemporalExtent			 
 SubClassOf: roleProvidedBy some owl:Thing 
          or rolePerformedBy some owl:Thing
Class: TemporalExtent DisjointWith: Role
ObjectProperty: hasTemporalExtent
ObjectProperty: roleProvidedBy  Domain: Role InverseOf: providesRole 
ObjectProperty: rolePerformedBy Domain: Role InverseOf: performsRole   
\end{lstlisting}
 %  \columnbreak
 %    \lstinputlisting[firstline=2]{TimeVaryingEntitities/FunctionWithInverseAndRange.dol}
 %    \lstinputlisting[firstline=2]{TimeVaryingEntitities/ClassPairOneMany.dol}
 % \end{multicols}
\vspace{-1em}    \caption{\GODP{Role} ODP from \cite{DBLP:books/ios/p/Krisnadhi16a} in OWL 2 Manchester syntax
     }
     \label{fig:RolePatterns}
 \end{figure}

\subsection{The Original Ontology Design Pattern for Role}

In \cite{DBLP:books/ios/p/Krisnadhi16a}, %two generic role patterns 
the ontology design pattern for Role is introduced,
see Fig.~\ref{fig:Role-diagram} for an overview and Fig.~\ref{fig:RolePatterns}
for its axiomatisation  (which we have converted to OWL 2 Manchester syntax).
To summarise, in \cite{DBLP:books/ios/p/Krisnadhi16a} a role is an entity that has some temporal extent; it is provided by at most one entity and it is performed by at most one entity; and either an provider or a performer does exist.\footnote{For brevity, we have omitted two range declarations that are redundant because the correponding inverse roles have domain declarations.}
%\ednote{The redundant range declarations could also be marked as \%implied instead!}

In \cite{DBLP:books/ios/p/Krisnadhi16a} ODPs are instantiated by subsumption. E.g., for an instantiation of \GODP{Role} for  \GODP{Agent} the  axioms in Figure~\ref{fig:AgentSubSumPattern1} are added to the axioms in  Figure~\ref{fig:RolePatterns} in \cite{DBLP:books/ios/p/Krisnadhi16a}  (modulo translation into Manchester Syntax). The additional axioms in  \GODP{Agent} introduce \class{Agent} as class and 
 \class{AgentRole} as a subclass of \class{Role} that is performed by agents. Lastly, it is asserted roles that are performed by agents are agent roles.  

 \begin{figure}[hbt]
% \begin{multicols}{2}
\begin{lstlisting}
Class: Agent 
Class: AgentRole  SubClassOf: Role, rolePerformedBy only Agent 
Class: RolePerformedBySomeAgent  
 EquivalentTo: rolePerformedBy some Agent  
 SubClassOf: AgentRole 
\end{lstlisting}
%  \end{multicols}
\vspace{-1em}
\caption{\GODP{Agent} defined as instantiation of \GODP{Role} via subsumption in  \cite{DBLP:books/ios/p/Krisnadhi16a} }
\label{fig:AgentSubSumPattern1}
 \end{figure}

Note that \GODP{Agent} may be considered as a new ODP, as done in \cite{DBLP:books/ios/p/Krisnadhi16a}. However, it is an instantiation of the \GODP{Role} pattern in the sense that Figure~\ref{fig:AgentSubSumPattern1} serves as illustration how \GODP{Role} is to be instantiated not just for \class{Agent}, but also in other circumstances. 
For example, if we would want to instantiate  \GODP{Role} for \class{Patient} (or some other thematic role), we presumably intended to use similar axioms. Thus, from our perspective the 
 ODP \GODP{Role} does not just consist of the  axioms in  Figure~\ref{fig:RolePatterns}, but also 
  includes the instructions on how to instantiate it in a given context. This instruction is  
 given here by the \class{Agent} example.

Illustrating how an ODP is supposed to be instantiated by providing an example is useful, but the technique has some signifiant drawbacks.  Firstly, this methodology requires the user 
to understand the ODP in detail. For example, without familiarity of the  \GODP{Role}, the necessity for adding the last two axioms in Figure~\ref{fig:AgentSubSumPattern1}  is not apparent.
Secondly, the example may not cover important aspects. For example,  in 
 the case of \GODP{Role} the example does not cover how one is supposed to instantiate the ODP for 
  roles that depend on institutions. Hence without detailed understanding of \GODP{Role} a user will not be able to use \GODP{Role} in order to, for example, represent that Bernd is a professor 
   at the University of Bremen. And given that there are no explicit rules on how to do so, different users may find different approaches to represent institutional roles as instantiations of \GODP{Role}. Thus, if our goal is to enable a team of ontology developers with a moderate level of ontology expertise to use ODPs in a way that is consistent across a large ontology, we need a better methodology than illustration by example.

\subsection{Generic Ontology Design Pattern for Role via Subsumption}
   \label{sec:subsumption}
GODPs uses \GDOL in order to explicitly encode how a pattern is supposed to be instantiated. There are two different strategies for the instantiation of a pattern, namely subsumption and parametrisation.\footnote{In programming language contexts, these are called subtype polymorphism and parametric polymorphism, resp.}  Since \cite{DBLP:books/ios/p/Krisnadhi16a} uses a subsumption strategy, we consider it first. 
\GODP{RoleGODPSubsumption} in Figure~\ref{fig:GODPSubsum} provides an axiomatisation that  follows faithfully  the \GODP{Agent} example in Figure~\ref{fig:AgentSubSumPattern1}.  
However, instead of \class{AgentRole} and \class{Agent}, the GODP uses two parameters, namely \class{RoleKind} and \class{Performer}. 
As we discussed in Section~\ref{sect:GenericOntologiesGDOL}, a GODP may be instantiated by providing appropriate arguments. 
In ontology \GODP{ThematicRoles} the GODP \GODP{RoleGODPSubsumption} is instantiated for three different thematic roles. Its definition of \class{Agent} and \class{AgentRole}
is equivalent to \GODP{Agent} as defined in Figure~\ref{fig:AgentSubSumPattern1}.   
As this example illustrates, 
one benefit of defining GODPs in \GDOL is the convenience of being able to instantiate a GODP without needing to repeatedly write axioms as in Figure~\ref{fig:AgentSubSumPattern1}. Indeed one is able to instantiate a GODP without considering the details of its axiomatisation. Further, it guarantees that a GODP is always instantiated in the same way and no axioms are forgotten.

 \begin{figure}[bth]
% \begin{multicols}{2}
\begin{lstlisting}[firstline=2]
pattern RoleGODPSubsumption [Class: RoleKind] [Class: Performer] =
 Role_original then 
 Class: RoleKind  SubClassOf: Role, rolePerformedBy only Performer 
 Class: RolePerformedBySome[Performer]  
  EquivalentTo: rolePerformedBy some Performer
  SubClassOf: RoleKind 
 
ontology ThematicRoles = 
    RoleGODPSubsumption[AgentRole][Agent]
and RoleGODPSubsumption[PatientRole][Patient]	
and RoleGODPSubsumption[Instrument][Instrument]
 
\end{lstlisting}
\vspace{-1em}
\caption{The \GODP{Role} GODP with instantiation via subsumption} 
\label{fig:GODPSubsum}
 \end{figure}

 Figure~\ref{fig:GODPSubsum} illustrates an important feature of \GDOL, namely parametrised names like  \class{RolePerformedBySome[Performer]}. 
Brackets ``\technical{[}'' and  ``\technical{]}''  are used around constituent names. %, and comma  ``\technical{,}''  as separator.
Such a bracketed construct in a name appears at its end, and may contain several constituents, separated by comma  ``\technical{,}''
As a result of substituting parameters by arguments in the expansion of an instantiation, the constituent names are substituted by the corresponding argument names, e.g. \class{RolePerformedBySome[Agent]}, where \class{Agent} is the respective Performer.
At the end of this flattening process, parameterised names are \emph{stratified}: all occurrences of ``\technical{[}'' or ``\technical{,}'' are turned into ``\technical{\_}'', and (trailing) occurrences of  ``\technical{]}'' are dropped. Thus, the stratified version of 
\GODP{ThematicRoles} contains three different classes, namely 
\class{RolePerformedBySome\_Agent}, 
\class{RolePerformedBySome\_Patient}, and
\class{RolePerformedBySome\_Instrument}.

If we would substitute the parametrised name \class{RolePerformedBySome[Performer]} by a `normal' class name, e.g., \class{RolePerformedBySomePerformer}, then \GODP{ThematicRoles} would only contain that class. In this case it would follow that 
\class{RolePerformedBy some Agent} is equivalent with \class{rolePerformedBy some Patient}. Hence, provided the other axioms, 
  $\class{Agent}(a)$ and 
$\relation{performedsRole}{(a,r)}$ would entail that  $\class{Patient}(a)$\footnote{If we additionally  assume that every agent and every patient performs  some agent role or patient
 role, respectively, then the axioms would even  entail that classes  \class{Agent} and \class{Patient} are equivalent.} 
 -- which is clearly not intended.
This example illustrates how the use of parametrised names enables the use of several different instantiations of one GODP within one larger ontology. 
   
Parametrised names are a seemingly innocuous feature, but it considerably reduces parameter lists and simplifies instantiations and re-use, since every instantiation automatically generates a new set of (stratified) names, cf. \cite{GODPs-WOP17}.

\subsection{Generic Ontology Design Pattern for Role via Parametrisation}
\label{sec:paramet}
As we have seen, instantiation of ODPs via subsumption follows the following strategy: one imports 
the pattern ontology and  extends it by adding subclass axioms. As a result the pattern ontology is 
 included in the final result, and all instantiations of the pattern are linked within the 
  subsumption hierarchy. For example, in the ontology ThematicRoles in Figure~\ref{fig:GODPSubsum} 
   the classes \class{AgentRole}, \class{PatientRole}, and \class{InstrumentRole} would all be 
    subclasses of \class{Role}. In this particular case this may be desirable, but this is not always the case. 
	
	For example, the 
\GODP{Role} ODP as axiomatised in \cite{DBLP:books/ios/p/Krisnadhi16a} (see 	Figure~\ref{fig:RolePatterns}) includes a recurrent pattern itself. The axioms for \relation{providesRole} and \relation{performsRole} are structurally identical: they are partial functions with inverses and their range is restricted to some class. 
Thus, we may want to identify this pattern as its own GODP, let's call it \GODP{ScopedPartialFunctionWithInverse}\footnote{The function is scoped in the sense that it does not apply to the whole domain of quantification, but restricted to instances of some class, e.g., \class{Role}.}. If we would instantiate this GODP with help of the subsumption strategy, the resulting ontology would  include an object property \relation{ScopedPartialFunctionWithInverse}  and axioms that establish
\relation{providesRole} and \relation{performsRole} as  subproperties of it. While these axioms are not false, it seems rather undesirable -- at least to the authors -- to be forced to include these kind of   axioms into ones ontology. This is because it clutters the ontology with symbols and axioms that are just technically motivated and are not specific to its domain.  We will return to this example 
  below, but for now we just conclude that under some circumstances the subsumption strategy is not ideal, because it requires the importation of the ODS into the final product. 
  The wholesale importation of the ODS is even more problematic, if the ODS contains some parts that are regarded as optional and are not necessarily useful for every user of the ODS.

 \begin{figure}[bth]
\begin{lstlisting}[firstline=2,lastline=18]
pattern RoleGODPParametrisation 
  [Class: Role] [Class: Performer] ?[Class: Provider] =

Class: Role 
 SubClassOf: roleProvidedBy[Provider] max 1 Provider, 
             rolePerformedBy[Performer] max 1 Performer,  
             hasTemporalExtent some TemporalExtent, 
             hasTemporalExtent only TemporalExtent 			 
 SubClassOf: roleProvidedBy[Provider] some Provider 
          or rolePerformedBy[Performer] some  Performer
Class: TemporalExtent
ObjectProperty: hasTemporalExtent
ObjectProperty: provides[Role]    Range:  Role 
ObjectProperty: roleProvidedBy[Provider]  InverseOf: provides[Role]             
ObjectProperty: performs[Role]    Range:  Role               
ObjectProperty: rolePerformedBy[Performer] InverseOf: performs[Role]   
DisjointClasses: Role, TemporalExtent
\end{lstlisting}
%\vspace{-1.em}\flushleft{[...]}
\vspace{-1em}
 \caption{The \GODP{Role} GODP with instantiation via parametrisation}
\label{fig:GODPPara}
 \end{figure}

The parametrisation strategy for the instantiation of ODPs solves both of these issues. As we have seen in 
 Section~\ref{sect:GenericOntologiesGDOL} the application of a parametrised ontology to some arguments leads to a new ontology, in which the parameters are substituted by their arguments. 
Hence, the generic terms in the GODP are no longer occurring in the resulting ontology. Furthermore, \GDOL supports optional parameters \cite{GDOL-WoMoCOE19}, which may be used to include optional parts of an ontology design pattern.

To see how both of these aspects of \GDOL are utilised,  let's return to the \GODP{Role} example from Figure~\ref{fig:Role-diagram}. However, let's assume that the left hand side  of the diagram (e.g., the role provider) is optional. 
In this case, we may define the \GODP{Role} GODP as in Figure~\ref{fig:GODPPara}.
%\footnote{For the sake of brevity we include only the relevant axioms. The GODP also axioms identical to lines 
%11-17 in Figure~\ref{fig:RolePatterns}.} 
This GODP has three parameters, namely \class{Role}, \class{Performer}, and \class{Provider}, the latter is marked as optional with a question mark. 
Because we need to distinguish between the performers and providers, the occurrences of 
  \class{owl:Thing} in \GODP{Role\_Original} in  Figure~\ref{fig:RolePatterns} have been replaced 
   by the appropriate classes. 
In this example we decided that the instantiations of the pattern should not use the generic object properties \relation{roleProvidedBy}, \relation{rolePerformedBy} and their inverses, but  rather generate specific instantiations of these. For this purpose we use a parametrised object  property marked by terms in square brackets (see also example in Section~\ref{sec:subsumption}).

% If we apply \GODP{RoleGODPParametrisation} to the arguments \class{Role}
% \class{owl:Thing} and \class{owl:Thing}, we get the similar axioms as in \GODP{Role\_Original}.
% More interestingly,
If we apply \GODP{RoleGODPParametrisation} to the arguments 
\class{ProfRole}, \class{Professor} and \class{University}, the flattened ontology does not contain either of the terms \class{Role}, \class{Performer}, or \class{Performer} because they are 
 substituted (see Figure~\ref{fig:ProfRole})\footnote{For the sake of brevity we include only the first axioms.}. 
  Hence, the resulting ontology is more compact than the ontology that would have been the result of the subsumption strategy. Furthermore, some of the symbols of the GODP have been replaced by 
   instance specific variants (e.g., \relation{roleProvidedBy\_University}) while others are unchanged (e.g., \relation{hasTemporalExtent}).

 \begin{figure}[t]
% \begin{multicols}{2}
\begin{lstlisting}
Class: ProfRole 
 SubClassOf: roleProvidedBy_University max 1 University, 
             rolePerformedBy_Professor max 1 Professor,
             hasTemporalExtent some TemporalExtent, 
             hasTemporalExtent only TemporalExtent 			 
 SubClassOf: roleProvidedBy_University some University 
          or rolePerformedBy_Professor some Professor
\end{lstlisting}
%  \end{multicols}
\vspace{-1em}
\caption{Result of \GODP{RoleGODPParametrisation[Class: ProfRole][Class: Professor][Class: University]}}
\label{fig:ProfRole}
 \end{figure}

In case an optional argument is omitted, all corresponding axioms are deleted. 
Figure~\ref{fig:MotherRole} shows the result of applying 
  \GODP{RoleGODPParametrisation} to the arguments \class{MotherRole} and  \class{Mother} with no third argument. Since no argument for the \class{Provider} parameter is given, the 
 lines 2 and 6-7 from Figure~\ref{fig:ProfRole} have no counterpart in Figure~\ref{fig:MotherRole}.
  
 This example illustrates that sometimes care needs to be taken with
 optional arguments. Some people would expect the axiom \emph{Class: MotherRole  SubClassOf: rolePerformedBy 
   some Mother} to be included among the axioms in Figure~\ref{fig:MotherRole}. However, since missing optional  arguments lead to the deletion of axioms, the whole axiom is removed. Removing only the first disjunct would strengthen the axiom. Hence, it is the user's decision whether this strengthening is useful
 or not.

 \begin{figure}[t]
% \begin{multicols}{2}
\begin{lstlisting}
Class: MotherRole
 SubClassOf: rolePerformedBy_Mother max 1 Mother,
             hasTemporalExtent some TemporalExtent, 
             hasTemporalExtent only TemporalExtent 			 
\end{lstlisting}
%  \end{multicols}
\vspace{-1em}
\caption{Result of \GODP{RoleGODPParametrisation[Class: MotherRole][Class: Mother][]}}
\label{fig:MotherRole}
 \end{figure}

\subsection{Patterns Within Patterns}

The presentation of the \GODP{Role} ODP in diagram in Fig.~\ref{fig:Role-diagram}
consists of three parts, namely:
1) the relation of \class{Role} to the provider of the role,
2) the relation of \class{Role} to the performer of the role, and
3) the relation of \class{Role} to \class{TemporalExtent}.
However, this three-part structure is lost in the  axiomatisation of \GODP{Role} in Fig.~\ref{fig:RolePatterns}. An axiomatisation that reflects the modular structure of the  
\GODP{Role} ODP has several benefits. Firstly, subdividing the axioms into meaningful modules may 
 improve the readability of the  ODP. Secondly, it enables the reuse of these modules as ODPs on 
  their own. For example, the axiomatisation of \class{TemporalExtent} seems not particular to 
   \class{Role}. Thus, it seems sensible to provide this part of the axiomatisation as its own independent ODP that may be reused in other contexts. Finally, the decomposition of  \GODP{Role} into smaller GODPs  helps to avoid code duplication. 
As we discussed in Section~\ref{sec:paramet}, the axioms about the provider and the performer are 
 structurally identical, since both introduce partial functions that are restricted to some class as   well their inverse. By introducing a GODP called \GODP{ScopedPartialFunctionWithInverse} we may avoid maintaining the axioms for this kind of concept twice.

Let's start with the \GODP{ScopedPartialFunctionWithInverse} GODP (see Figure~\ref{fig:GODPartialFunction}).     
The name of the GODP concisely expresses
the involved mathematical properties: a relation with inverse that is
a partial function, not necessarily on the whole of \class{owl:Thing}, but on a class \class{D}  (and, thus, {scoped}). 
In \cite{DBLP:books/ios/p/Krisnadhi16a} the ranges of \relation{rolePerformedBy} and \relation{roleProvidedBy} are not restricted, see  \class{owl:Thing} (in Figure~\ref{fig:Role-diagram}). Since we intend to enable a distinction between the providers and performers of roles, we include also a range argument \class{R} for  \GODP{ScopedPartialFunctionWithInverse}.

 \begin{figure}[bt]
\begin{lstlisting}[firstline=2]
pattern ScopedPartialFunctionWithInverse 
 [ObjectProperty: f][Class: D][Class: R] [ObjectProperty: finv] =
ObjectProperty: f Domain: D  Range: R  
ObjectProperty: finv   InverseOf: f
Class: D SubClassOf: f max 1 R
\end{lstlisting}
\vspace{-1em}
\begin{lstlisting}[firstline=2]
pattern HasTemporalExtent [Class: C] =
Class: C SubClassOf: hasTemporalExtent some TemporalExtent, 
                     hasTemporalExtent only TemporalExtent
Class: TemporalExtent DisjointWith: C
ObjectProperty: hasTemporalExtent
\end{lstlisting}
\vspace{-1em}
 \caption{GODPs \GODP{ScopedPartialFunctionWithInverse} and \GODP{hasTemporalExtent}}
\label{fig:GODPartialFunction}
 \end{figure}

The GODP \GODP{hasTemporalExtent} asserts that the instances of a given class \class{C} have some temporal extent, and that \class{C} is disjoint with \class{TemporalExtent}. In our running example
these axioms are asserted about \class{Role}, but they hold for a wide range of classes and, thus, may be reused in other contexts.

With these two building blocks we may now restructure \GODP{RoleGODPParametrisation} along the lines of the diagram in Figure~\ref{fig:Role-diagram}. Figure~\ref{fig:GODPdecomposed} shows the resulting 
 GODP \GODP{RoleGODPDecomposed}. It uses the same parameters as \GODP{RoleGODPParametrisation}.  
  However, its body is significantly shorter, since most axioms have been replaced by  applications of GODPs. In line 3 it is asserted that the instances of the class that is provided as argument for the parameter \class{Role} have temporal extent. 
 Lines 4-5 assert that the variant of \relation{rolePerformedBy} that is generated depending on the argument for \relation{Performer} is a scoped partial function (and has an appropriate inverse). 
 Lines 6-7 assert  the same for \relation{roleProvidedBy}. These three applications of GODPs correspond the three parts of the diagram in Figure~\ref{fig:Role-diagram}. The last axiom is both about \class{Provider} and \class{Performer}. Hence, it is not part of either sub-GOPD of \GODP{RoleGODPDecomposed}, but the axiom is added in order to connect the different parts of the GODP.
 \begin{figure}[bt]
\begin{lstlisting}[firstline=2]
pattern RoleGODPDecomposed 
  [Class: Provider] [Class: Role] ?[Class: Performer] =
HasTemporalExtent [Role] then
ScopedPartialFunctionWithInverse [rolePerformedBy[Performer]] 
                          [Role] [Performer] [performs[Role]] then  
ScopedPartialFunctionWithInverse [roleProvidedBy[Provider]]
                          [Role] [Provider]  [provides[Role]]   then 
Class: Role SubClassOf: roleProvidedBy[Provider]   some Performer 
                     or rolePerformedBy[Performer] some Provider
\end{lstlisting}
 \vspace{-1em}
 \caption{\GODP{RoleGODPDecomposed} consists of a decomposition of the GODP in 3 subparts}
\label{fig:GODPdecomposed}
 \end{figure}

\section{Conclusion, Related and Future Work}
\label{sect:FutureWork}

%%%%%%%%%%%%%%%%%%%%%%%%%%
We have shown that through the use of generic ontology design patterns
(GODPs), one can ease and inrease reuse of ontologies as well as avoid
code duplication, resulting in enhanced quality of ontologies. In
contrast to informal ontology design patterns, which often just are
ontologies, GODPs have a clear instantiation semantics (implemented by
\Hets) that allows their tailoring for specific context.

The OTTR approach of \technicalterm{Parameterized Templates} with
macro expansion \cite{OTTR-WOP17,OTTR-ISWC18} is in many respects very
similar to the GODP approach.
%Apart from the benefit of being able to
%state semantic preconditions using ontologies as parameters (not
%elaborated here, cf. \cite{GODPs-WOP17}),
The major
contribution of the {{\GenericDOL}} approach described in this paper
is parametrisation, supplemented by parameterized names. This is crucial
for avoiding unintended name clashes. An
extra benefit are the other structuring operations of DOL and the
smooth integration with heterogeneous modelling. % using DOL and Hets.
In particular, this means that GODPs can be used with other ontology languages
than OWL2, for example, also with Common Logic. This is not possible with OTTR.

%% In \cite{GODPs-WOP17} we have shown how a complex pattern can
%% be structured into sub-patterns, and how iteration can be achieved.
%% An extension of {\GenericDOL} for optional parameters (as already used
%% here), parameter lists and general features for iteration has been
%% proposed in \cite{GDOL-WoMoCOE19}.  A GODP will then e.g. allow a
%% compact solution for an inheritance of all induced properties given in
%% a list. % (cf. Sect.~\ref{sect:InducedProperties}.

In this paper, we have concentrated on one example from the
literature. We expect that salient and well-known patterns in the
existing repositories like \url{ontologydesignpatterns.org}, primarily
conceptual or ``knowledge'' ODPs, will soon be cast into the GODP (or
OTTR) framework to make reuse more practical.

All ontologies in this paper can be found online in the Ontohub
repository \url{http://www.ontohub.org/godp},
available both through git and through a web interface.
\vspace{-1.5em}

%put back in for final version
\section*{Acknowledgements}
\label{sec:Acknowledgements}
We are very grateful to
%the reviewers and
Serge Autexier,
Mihai Codescu, %Fabian Neuhaus,
Jens Pelzetter, and Martin Rink
for their suggestions and contributions.

\nocite{WOP17}

%\section*{References}
%\bibliographystyle{elsarticle-harv}
%\bibliographystyle{splncs}
\bibliographystyle{plain}
\bibliography{ODLSJowo}

\end{document}